\pdfoutput=1
\documentclass[11pt]{article}
\usepackage{CJKutf8}
\usepackage{CJK}
\usepackage{threeparttable}
\usepackage{graphicx}
\usepackage{float} 
\usepackage{subfigure}
\usepackage{caption}
\usepackage[]{ACL2023}

\usepackage{times}
\usepackage{latexsym}
\usepackage[T1]{fontenc}
\usepackage[utf8]{inputenc}

\usepackage{microtype}

\usepackage{inconsolata}
\usepackage{bm}
\usepackage{amsfonts}
\usepackage{amsmath}
\usepackage{multirow}
\usepackage[]{placeins}
\allowdisplaybreaks[4]
\title{Exploring Better Text Image Translation with Multimodal Codebook}

\author{
    Zhibin Lan\textsuperscript{1,3}\footnotemark[1], 
    Jiawei Yu\textsuperscript{1,3}\thanks{~~Equal contribution.}, 
    Xiang Li\textsuperscript{2}, 
    Wen Zhang\textsuperscript{2}, 
    Jian Luan\textsuperscript{2}\\ 
    \textbf{Bin Wang\textsuperscript{2}, 
    Degen Huang\textsuperscript{4}, 
    Jinsong Su\textsuperscript{1,3}\thanks{~~Corresponding author.}}\\
    \textsuperscript{1}School of Informatics, Xiamen University, China\\
    \textsuperscript{2}Xiaomi AI Lab, Beijing, China\\
    \textsuperscript{3}Key Laboratory of Digital Protection and Intelligent Processing of Intangible Cultural Heritage \\
    of Fujian and Taiwan (Xiamen University), Ministry of Culture and Tourism, China\\
    \textsuperscript{4}Dalian University of Technology, China\\
    \texttt{\small\{lanzhibin,yujiawei\}@stu.xmu.edu.cn
    }~~
    \texttt{\small jssu@xmu.edu.cn
    }
}
\begin{document}

\maketitle

\begin{abstract}
Text image translation (TIT) aims to translate the source texts embedded in the image to target translations, which has a wide range of applications and thus has important research value. However, current studies on TIT are confronted with two main bottlenecks: 1) this task lacks a publicly available TIT dataset, 2) dominant models are constructed in a cascaded manner, which tends to suffer from the error propagation of optical character recognition (OCR). In this work, we first annotate a Chinese-English TIT dataset named OCRMT30K, providing convenience for subsequent studies. Then, we propose a TIT model with a multimodal codebook, which is able to associate the image with relevant texts, providing useful supplementary information for translation. Moreover, we present a multi-stage training framework involving text machine translation, image-text alignment, and TIT tasks, which fully exploits additional bilingual texts, OCR dataset and our OCRMT30K dataset to train our model. Extensive experiments and in-depth analyses strongly demonstrate the effectiveness of our proposed model and training framework.\footnote{Our code and dataset can be found at \url{https://github.com/DeepLearnXMU/mc_tit}}
\end{abstract}

\section{Introduction}
In recent years, multimodal machine translation (MMT) has achieved great progress and thus received increasing attention. Current studies on MMT mainly focus on the text machine translation with scene images \cite{DBLP:conf/acl/ElliottFSS16,DBLP:conf/acl/CalixtoLC17,DBLP:conf/ijcnlp/ElliottK17,DBLP:conf/wmt/LibovickyHM18,DBLP:conf/acl/Distilling,DBLP:conf/iclr/uvrnmt, DBLP:journals/mt/SulubacakCGREST20}. However, a more common requirement for MMT in real-world applications is text image translation (TIT) \cite{DBLP:conf/icpr/MaZTHWZ022}, which aims to translate the source texts embedded in the image to target translations. Due to its wide applications, the industry has developed multiple services to support this task, such as Google Camera Translation.

\begin{figure}[]
    \centering
    \setlength{\abovecaptionskip}{2pt}
    \includegraphics[width=\linewidth]{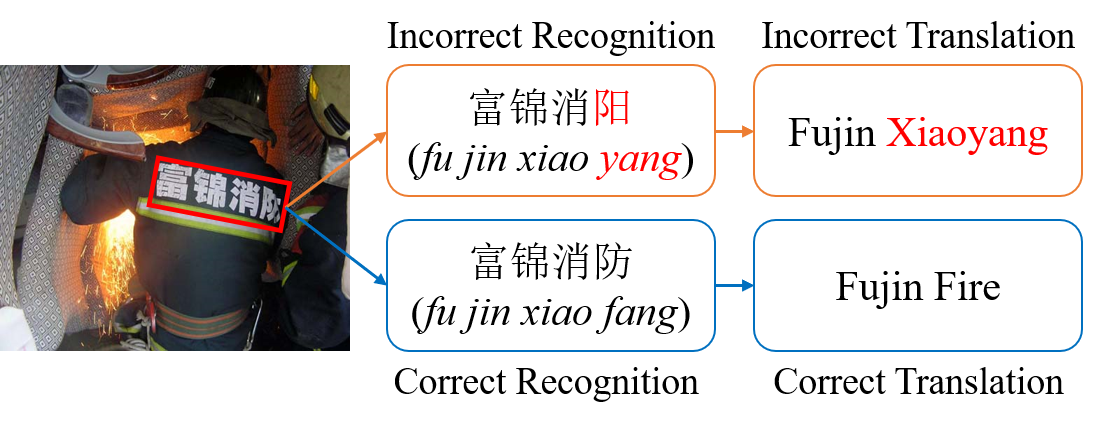}
    \caption{An example of text image translation. The Bounding box in red represents the text to be recognized. We can observe that the incorrect OCR result will negatively affect the subsequent translation.}
    \label{figure:1}
\end{figure}

Current studies on TIT face two main bottlenecks. First, this task lacks a publicly available TIT dataset. Second, the common practice is to adopt a cascaded translation system, where the texts embedded in the input image are firstly recognized by an optical character recognition (OCR) model, and then the recognition results are fed into a text-only neural machine translation (NMT) model for translation. However, such a method tends to suffer from the problem of OCR error propagation, and thus often generates unsatisfactory translations. As shown in Figure \ref{figure:1}, \begin{CJK}{UTF8}{gbsn}“\textit{富锦消防}”\end{CJK} ("\textit{fu jin xiao fang}”) in the image is incorrectly recognized as \begin{CJK}{UTF8}{gbsn}“\textit{富锦消阳}”\end{CJK} (“\textit{fu jin xiao yang}”). Consequently, the text-only NMT model incorrectly translates it into “\textit{Fujin Xiaoyang}”. Furthermore, we use the commonly-used PaddleOCR\footnote{https://github.com/PaddlePaddle/PaddleOCR.} to handle several OCR benchmark datasets. As reported in Table \ref{table:1}, we observe that the highest recognition accuracy at the image level is less than 67\% and that at the sentence level is not higher than 81\%. It can be said that OCR errors are very common, thus they have a serious negative impact on subsequent translation.

\begin{table}[]
\centering
\resizebox{0.9\columnwidth}{!}{
\begin{threeparttable} %添加此处
\begin{tabular}{c|c|c}
\hline
\textbf{Dataset}   & \begin{tabular}[c]{@{}c@{}}\textbf{Image Level} \\ \textbf{Accuracy}\end{tabular} & \begin{tabular}[c]{@{}c@{}}\textbf{Sentence Level} \\ \textbf{Accuracy}\end{tabular} \\ \hline
RCTW-17      & 65.27\%                                                           & 80.20\%                                                              \\
CASIA-10K & 43.63\%                                                           & 69.79\%                                                              \\
ICDAR19-ArT & 50.96\%                                                           & 75.84\%                                                              \\
ICDAR19-MLT & 66.63\%                                                           & 80.77\%  
                                                       \\
ICDAR19-LSVT & 43.97\%                                                           & 75.70\%  
                                                        \\ \hline
\end{tabular}
\end{threeparttable} %添加此处
}
\caption{Recognition accuracies on five commonly-used OCR datasets. Image level accuracy refers to the proportion of correct recognitions among all images. Sentence level accuracy denotes the proportion of correctly recognized sentences among all recognized sentences.}
% \vspace{-0.3cm}
\label{table:1}
\end{table}

In this paper, we first manually annotate a Chinese-English TIT dataset named OCRMT30K, providing convenience for subsequent studies. This dataset is developed based on five Chinese OCR datasets, including about 30,000 image-text pairs. 

Besides, we propose a TIT model with a multimodal codebook to alleviate the OCR error propagation problem. The basic intuition behind our model is that when humans observe the incorrectly recognized text in an image, they can still associate the image with relevant or correct texts, which can provide useful supplementary information for translation. Figure \ref{figure:2} shows the basic architecture of our model, which mainly consists of four modules: 1) a \textit{text encoder} that converts the input text into a hidden state sequence; 2) an \textit{image encoder} encoding the input image as a visual vector sequence; 3) a \textit{multimodal codebook}. This module can be described as a vocabulary comprising latent codes, each of which represents a cluster. It is trained to map the input images and ground-truth texts into the shared semantic space of latent codes. During inference, this module is fed with the input image and then outputs latent codes containing the text information related to ground-truth texts. 4) a \textit{text decoder} that is fed with the combined representation of the recognized text and the outputted latent codes, and then generates the final translation.

Moreover, we propose a multi-stage training framework for our TIT model, which can fully exploit additional bilingual texts and OCR data for model training. Specifically, our framework consists of four stages. \textit{First}, we use a large-scale bilingual corpus to pretrain the text encoder and text decoder. \textit{Second}, we pretrain the newly added multimodal codebook on a large-scale monolingual corpus. \textit{Third}, we further introduce an image encoder that includes a pretrained vision Transformer with fixed parameters to extract visual features, and continue to train the multimodal codebook. Additionally, we introduce an image-text alignment task to enhance the ability of the multimodal codebook in associating images with related texts. \textit{Finally}, we finetune the entire model on the OCRMT30K dataset. Particularly, we maintain the image-text alignment task at this stage to reduce the gap between the third and fourth training stages.

Our main contributions are as follows:
\begin{itemize}
% \vspace {-\topsep}
% \setlength{\itemsep}{0pt}
% \setlength{\parsep}{0pt}
% \setlength{\parskip}{0pt}
\item We release an OCRMT30K dataset, which is the first Chinese-English TIT dataset, prompting the subsequent studies.
\item We present a TIT model with a multimodal codebook, which can leverage the input image to generate the information of relevant or correct texts, providing useful information for the subsequent translation.
\item We propose a multi-stage training framework for our model, which effectively leverages additional bilingual texts and OCR data to enhance the model training.
\item Extensive experiments and analyses demonstrate the effectiveness of our model and training framework.
\end{itemize}

\section{Related Work}
In MMT, most early attempts exploit visual context via attention mechanisms \cite{DBLP:journals/corr/CaglayanBB16,DBLP:conf/wmt/HuangLSOD16,DBLP:conf/acl/CalixtoLC17,DBLP:conf/acl/LibovickyH17,DBLP:conf/emnlp/Incorporating,DBLP:journals/isci/SuCJZLGWL21}. Afterwards, \citeauthor{DBLP:conf/acl/Distilling} (\citeyear{DBLP:conf/acl/Distilling}) employ a translate-and-refine approach to improve translation drafts with visual context. Meanwhile, \citeauthor{DBLP:conf/acl/Latent} (\citeyear{DBLP:conf/acl/Latent}) incorporate visual context into MMT model through latent variables. Different from these studies focusing on coarse-grained visual-text alignment information, \citeauthor{DBLP:conf/acl/YinMSZYZL20} (\citeyear{DBLP:conf/acl/YinMSZYZL20}) propose a unified multimodal graph based encoder to capture various semantic relationships between tokens and visual objects. \citeauthor{DBLP:conf/mm/LinMSYYGZL20} (\citeyear{DBLP:conf/mm/LinMSYYGZL20}) present a dynamic context-guided capsule network to effectively capture visual features at different granularities for MMT.
% \citeauthor{DBLP:journals/isci/SuCJZLGWL21} (\citeyear{DBLP:journals/isci/SuCJZLGWL21}) su .

Obviously, the effectiveness of conventional MMT heavily relies on the availability of bilingual texts with images, which restricts its wide applicability. To address this issue, \citeauthor{DBLP:conf/iclr/uvrnmt} (\citeyear{DBLP:conf/iclr/uvrnmt}) first build a token-image lookup table from an image-text dataset, and then retrieve images matching the source keywords to benefit the predictions of target translation. Recently, \citeauthor{DBLP:conf/acl/FangF22} (\citeyear{DBLP:conf/acl/FangF22}) present a phrase-level retrieval-based method that learns visual information from the pairs of source phrases and grounded regions.

Besides, researchers investigate whether visual information is really useful for machine translation. \citeauthor{DBLP:conf/emnlp/Elliott18} (\citeyear{DBLP:conf/emnlp/Elliott18}) finds that irrelevant images have little impact on translation quality. \citeauthor{DBLP:conf/acl/WuKBLK20} (\citeyear{DBLP:conf/acl/WuKBLK20}) attribute the gain of MMT to the regularization effect. Unlike these conclusions, \citeauthor{DBLP:conf/naacl/CaglayanMSB19} (\citeyear{DBLP:conf/naacl/CaglayanMSB19}) and \citeauthor{DBLP:conf/emnlp/LiAS21} (\citeyear{DBLP:conf/emnlp/LiAS21}) observe that MMT models rely more on images when textual ambiguity is high or textual information is insufficient.

To break the limitation that MMT requires sentence-image pairs during inference, researchers introduce different modules, such as image prediction decoder \cite{DBLP:conf/ijcnlp/ElliottK17}, generative imagination network \cite{DBLP:conf/naacl/LongWL21}, autoregressive hallucination Transformer \cite{DBLP:conf/cvpr/LiPKCFCV22}, to produce a visual vector sequence that is associated with the input sentence.

Significantly different from the above studies on MMT with scene images, several works also explore different directions in MMT. For instance,  \citeauthor{DBLP:conf/eacl/MatusovWCSLC17} (\citeyear{DBLP:conf/eacl/MatusovWCSLC17}) and \citeauthor{DBLP:conf/mm/0003CJLXH21} (\citeyear{DBLP:conf/mm/0003CJLXH21}) investigate product-oriented machine translation, and other researchers focus on multimodal simultaneous machine translation \cite{DBLP:conf/emnlp/simultaneous20,DBLP:conf/eacl/IveLMCMS21}. Moreover, there is a growing body of studies on video-guided machine translation \cite{DBLP:conf/iccv/WangWCLWW19,DBLP:conf/acl/video21,DBLP:liyan}. These studies demonstrate the diverse applications and potential of MMT beyond scene images.

In this work, we mainly focus on TIT, which suffers from incorrectly recognized text information and is more practicable in real scenarios. The most related work to ours mainly includes \cite{Towards-end-to-end,Jain, DBLP:conf/icpr/MaZTHWZ022}. \citeauthor{Towards-end-to-end} (\citeyear{Towards-end-to-end}) first explore in-image translation task, which transforms an image containing the source text into an image with the target translation. They not only build a synthetic in-image translation dataset but also put forward an end-to-end model combining a self-attention encoder with two convolutional encoders and a convolutional decoder. \citeauthor{Jain} (\citeyear{Jain}) focus on the TIT task, and propose to combine OCR and NMT into an end-to-end model with a convolutional encoder and an autoregressive Transformer decoder. Along this line, \citeauthor{DBLP:conf/icpr/MaZTHWZ022} (\citeyear{DBLP:conf/icpr/MaZTHWZ022}) apply multi-task learning to this task, where MT, TIT, and OCR are jointly trained. However, these studies only center around synthetic TIT datasets, which are far from the real scenario.

\begin{figure}[]
\setlength{\abovecaptionskip}{2pt}
    \centering
    \includegraphics[width=\linewidth]{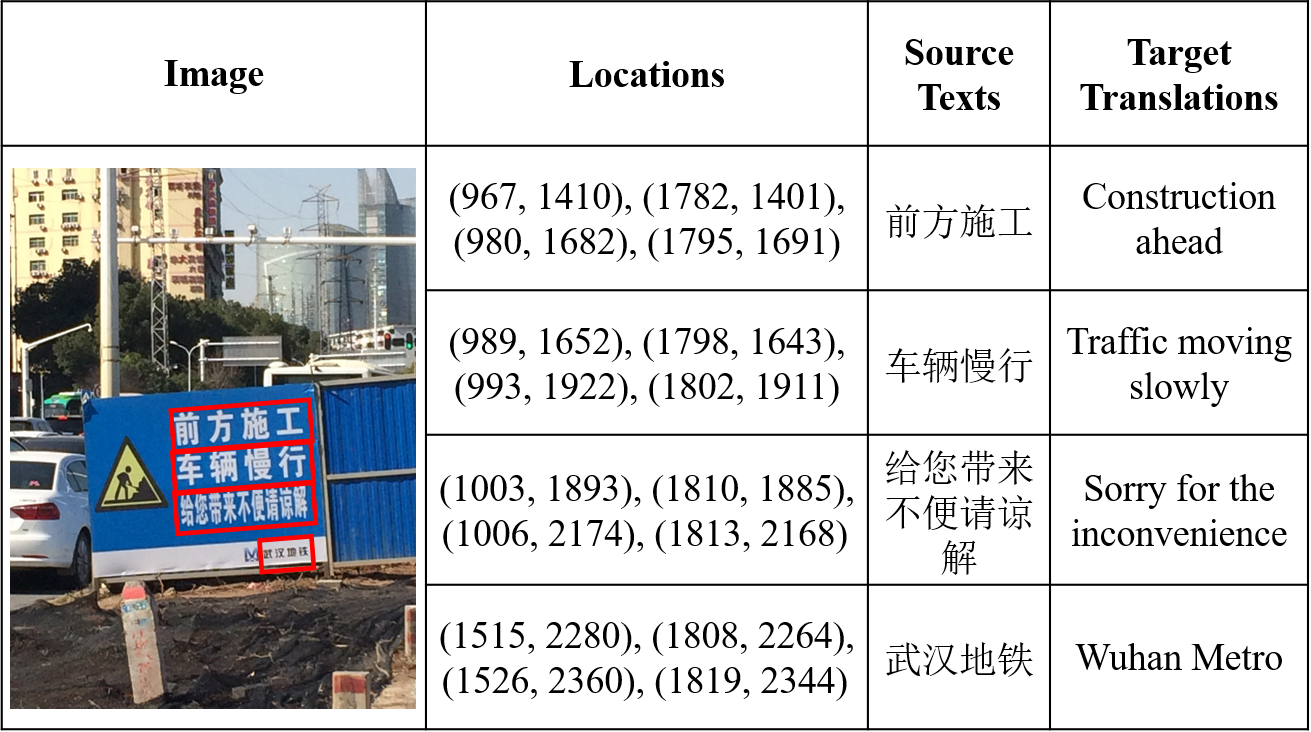}
    \caption{An example of the OCRMT30K dataset. The Locations are annotated by drawing bounding boxes to surround every text line.}
    \label{figure:example}
    % \vspace{-0.3cm}
\end{figure} 

\section{Dataset and Annotation}
To the best of our knowledge, there is no publicly available dataset for the task of TIT. Thus we first manually annotate a Chinese-English TIT dataset named OCRMT30K, which is based on five commonly-used Chinese OCR datasets: RCTW-17 \cite{DBLP:conf/icdar/ShiYLYXCBLB17}, CASIA-10K \cite{DBLP:journals/tip/HeZYL18}, ICDAR19-MLT \cite{DBLP:conf/icdar/NayefLOPBCKKM0B19}, ICDAR19-LSVT \cite{DBLP:conf/iccv/SunLLHDL19} and ICDAR19-ArT \cite{DBLP:conf/icdar/ChngDLKCJLSNLNF19}. We hire eight professional translators for annotation over five months and each translator is responsible for annotating 25 images per day to prevent fatigue. Translators are shown an image with several Chinese texts and are required to produce correct and fluent translations for them in English. In addition, we hire a professional translator to sample and check the annotated instances for quality control. We totally annotate 30,186 instances and the number of parallel sentence pairs is 164,674. Figure \ref{figure:example} presents an example of our dataset. 

\section{Our Model}
\subsection{Task Formulation}
In this work, following common practices \cite{DBLP:conf/lt4dh/AfliW16,DBLP:conf/icpr/MaZTHWZ022}, we first use an OCR model to recognize texts from the input image $\mathbf{v}$. Then, we fed both $\mathbf{v}$ and each recognized text $\hat{\mathbf{x}}$ into our TIT model, producing the target translation $\mathbf{y}$. In addition, $\mathbf{x}$ is used to denote the ground-truth text of $\hat{\mathbf{x}}$ recognized from $\mathbf{v}$.

To train our TIT model, we will focus on establishing the following conditional predictive probability distribution:
{
\begin{equation}
P(\mathbf{y}|\mathbf{v},\hat{\mathbf{x}};\bm{\theta})=\prod_{t=1}^{|\mathbf{y}|}P(y_t|\mathbf{v},\hat{\mathbf{x}},\mathbf{y}_{<t};\bm{\theta}),
\end{equation}}
where $\bm{\theta}$ denotes the model parameters.

\begin{figure}[]
\setlength{\abovecaptionskip}{5pt}
    \centering
    \includegraphics[width=1\linewidth]{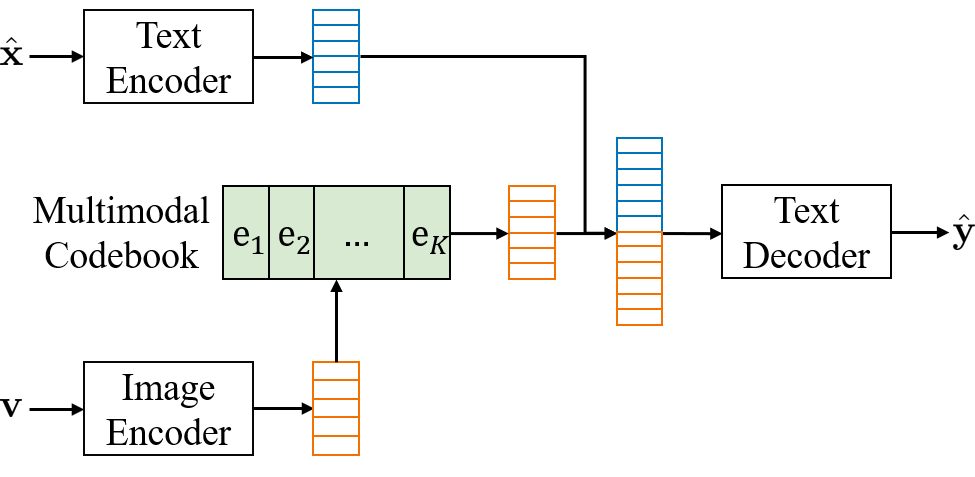}
    \caption{The overall architecture of our model includes a text encoder, an image encoder, a multimodal codebook, and a text decoder. Particularly, the multimodal codebook is the most critical module, which can associate images with relevant or correct texts. $\hat{\mathbf{x}}$ is the recognized text from the input image $\mathbf{v}$, $e_k$ represents the $k$-th latent code embedding and $\hat{\mathbf{y}}$ is the outputted target translation.}
    \label{figure:2}
\end{figure} 
% \vspace{-0.3cm}
\subsection{Model Architecture}

As shown in Figure \ref{figure:2}, our model includes four modules: 1) a \textit{text encoder} converting the input text into a hidden state sequence; 2) an \textit{image encoder} encoding the input image as a visual vector sequence; 3) a \textit{multimodal codebook} that is fed with the image representation and then outputs latent codes containing the text information related to the ground-truth text; and 4) a \textit{text decoder} that generates the final translation under the semantic guides of text encoder hidden states and outputted latent codes. All these modules will be elaborated in the following.

\textbf{Text Encoder}. Similar to dominant NMT models, our text encoder is based on the Transformer \cite{DBLP:conf/nips/VaswaniSPUJGKP17} encoder. It stacks $L_{e}$ identical layers, each of which contains a self-attention sub-layer and a feed-forward network (FFN) sub-layer.  

Let $\mathbf{H}_{e}^{(l)}=h_{e,1}^{(l)},h_{e,2}^{(l)},...,h_{e,N_e}^{(l)}$ denotes the hidden states of the $l$-th encoder layer, where $N_e$ is the length of the hidden states $\mathbf{H}_{e}^{(l)}$. Formally, $\mathbf{H}_{e}^{(l)}$ is calculated in the following way:
\begin{equation}
\mathbf{H}_{e}^{(l)}=\mathrm{FFN}(\mathrm{MHA}(\mathbf{H}_{e}^{(l-1)},\mathbf{H}_{e}^{(l-1)},\mathbf{H}_{e}^{(l-1)})),
\end{equation}

\noindent where $\mathrm{MHA}(\cdot,\cdot,\cdot)$ denotes a multi-head attention function \cite{DBLP:conf/nips/VaswaniSPUJGKP17}. Particularly, $\mathbf{H}_{e}^{(0)}$ is the sum of word embeddings and position embeddings. Note that we follow \citeauthor{DBLP:conf/nips/VaswaniSPUJGKP17} (\citeyear{DBLP:conf/nips/VaswaniSPUJGKP17}) to
use residual connection and layer normalization (LN) in each sub-layer, of which descriptions are omitted for simplicity. During training, the text encoder is utilized to encode both the ground-truth text $\mathbf{x}$ and the recognized text $\hat{\mathbf{x}}$, so we use $\mathbf{\hat{H}}_{e}^{(l)}$ to denote the hidden state of recognized text for clarity. In contrast, during inference, the text encoder only encodes the recognized text $\hat{\mathbf{x}}$, refer to Section \ref{sec:4.3} for more details.

\begin{figure*}[]
\setlength{\abovecaptionskip}{-2pt}

    \centering
    \includegraphics[width=0.95\linewidth]{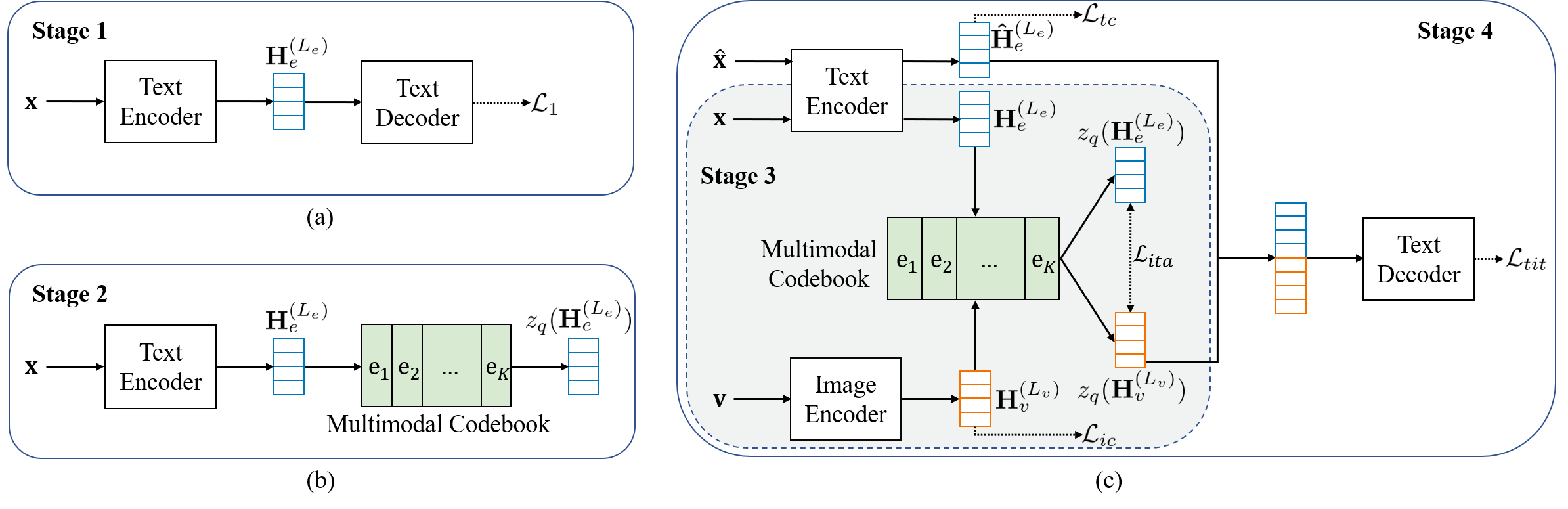}
    \label{figure:3c}
% \begin{minipage}[]{.5\linewidth}
%     \centering
%     % \captionsetup[subfigure]{labelformat=empty}
%     \subfigure{\label{figure:3c}\includegraphics[width=1.8\linewidth]{images/stage_all.png}}
% \end{minipage}

\caption{Overview of our multi-stage training framework. (a) Stage 1: we pretrain the text encoder and text decoder with $\mathcal{L}_{1}$. (b)
Stage 2: we update the multimodal codebook with an exponential moving average (EMA). (c) In Stage 3, we only train the dashed part of the model with $\mathcal{L}_{ita}$ and $\mathcal{L}_{ic}$. As for Stage 4, we optimize the whole model through $\mathcal{L}_{ita}$, $\mathcal{L}_{ic}$, $\mathcal{L}_{tit}$, and $\mathcal{L}_{tc}$.}
\label{figure:3}
% \vspace{-0.3cm}
\end{figure*}

\textbf{Image Encoder}. As a common practice, we use ViT \cite{DBLP:conf/iclr/DosovitskiyB0WZ21} to construct our image encoder. Similar to the Transformer encoder, ViT also consists of $L_v$ stacked layers, each of which includes a self-attention sub-layer and an FFN sub-layer. One key difference between the Transformer encoder and ViT is the placement of LN, where pre-norm is applied in ViT.

Given the image input $\mathbf{v}$, the visual vector sequence $\mathbf{H}_{v}^{(L_v)}=h_{v,1}^{(L_v)},h_{v,2}^{(L_v)},...,h_{v,N_v}^{(L_v)}$ output by the image encoder can be formulated as
\begin{equation}
    \mathbf{H}_{v}^{(L_v)} = \mathrm{MHA}(\mathbf{H}_e^{(L_e)}, \mathbf{W}_{v}  \mathrm{ViT}(\mathbf{v}),\mathbf{W}_v  \mathrm{ViT}(\mathbf{v})),
\end{equation}

\noindent where $N_v$ is the length of the hidden states $\mathbf{H}_{v}^{(L_v)}$ and $\mathbf{W}_v$ is a projection matrix to convert the dimension of $\mathrm{ViT}(\mathbf{v})$ into that of $\mathbf{H}_{e}^{(L_e)}$.

\textbf{Multimodal Codebook}. 
% It is a core component of our model, which can be viewed as a fixed-size table of embedding vectors. We denote it as $E=\{ e_1,e_2,...,e_K \} {\in} R^{K\times d}$ where $K$ is the size of the multimodal codebook, and $d$ is the dimension of latent code. In this work, we always set the dimension of the multimodal codebook to be the same as the hidden state of the text encoder. 
It is the core module of our model. The multimodal codebook is essentially a vocabulary with $K$ latent codes, each of which is represented by a $d$-dimensional vector $e_k$ like word embeddings. Note that we always set the dimension of the latent code equal to that of the text encoder, so as to facilitate the subsequent calculation in Equation \ref{codebook_update}.

With the multimodal codebook, we can quantize the hidden state sequence $\mathbf{H}_{e}^{(L_e)}=h_{e,1}^{(L_e)},h_{e,2}^{(L_e)},...,h_{e,N_e}^{(L_e)}$ or the visual vector sequence $\mathbf{H}_{v}^{(L_v)}=h_{v,1}^{(L_v)},h_{v,2}^{(L_v)},...,h_{v,N_v}^{(L_v)}$ to latent codes via a quantizer $z_q(\cdot)$.
Formally, the quantizer looks up the nearest latent code for each input, as shown in the following:
\begin{equation}
    z_q(h_{e,i}^{(L_e)}) =\mathop{\mathrm{argmin}}\limits_{e_{k^{'}}}||h_{e,i}^{(L_e)}-e_{k^{'}}||_2,
    \label{eq:5}
\end{equation}
\begin{equation}
    z_q(h_{v,j}^{(L_v)}) =\mathop{\mathrm{argmin}}\limits_{e_{k^{''}}}||h_{v,j}^{(L_v)}-e_{k^{''}}||_2.
\end{equation}

By doing so, both text and image representations are mapped into the shared semantic space of latent codes.

\textbf{Text Decoder}. This decoder is also based on the Transformer decoder, with $L_d$ identical layers. In addition to self-attention and FFN sub-layers, each decoder layer is equipped with a cross-attention sub-layer to exploit recognized text hidden states $\mathbf{\hat{H}}_{e}^{(L_e)}$ and latent codes $z_q(\mathbf{H}_v^{(L_v)})$.  

% Let $\mathbf{H}_{d}^{(l)}=h_{d,1}^{(l)},h_{d,2}^{(l)},...,h_{d,N_d}^{(l)}$ represents the hidden states of the $l$-th decoder layer containing $N_d$ hidden states, it is calculated using the following equations:
The hidden states of the $l$-th decoder layer are denoted by $\mathbf{H}_{d}^{(l)}=h_{d,1}^{(l)},h_{d,2}^{(l)},...,h_{d,N_d}^{(l)}$, where $N_d$ represents the total number of hidden states. These hidden states are calculated using the following equations:
\begin{equation}
\mathbf{C}_{d}^{(l)}=\mathrm{MHA}(\mathbf{H}_{d}^{(l-1)},\mathbf{H}_{d}^{(l-1)},\mathbf{H}_{d}^{(l-1)}),
\end{equation}
\begin{equation}
\mathbf{T}_{d}^{(l)}=[\mathbf{\hat{H}}_{e}^{(L_e)};z_q(\mathbf{H}_v^{(L_v)})],
\label{eq:concat}
\end{equation}
\begin{equation}
\mathbf{H}_{d}^{(l)}=\mathrm{FFN}({\mathrm{MHA}(\mathbf{C}_{d}^{(l)},
\mathbf{T}_{d}^{(l)},\mathbf{T}_{d}^{(l)}))}.
\end{equation}

Finally, at each decoding timestep $t$, the probability distribution of generating the next target token $y_t$ is defined as follows:
\begin{equation}
P(y_t|\mathbf{v},\hat{\mathbf{x}},\mathbf{y}_{<t};\bm{\theta})=\mathrm{softmax}(\mathbf{W}_o h_{d,t}^{(L_d)} + {b}_o),
\end{equation}
\noindent where $\mathbf{W}_o$ and ${b}_o$ are trainable model parameters.

\subsection{Multi-stage Training Framework}
\label{sec:4.3}
In this section, we present in detail the procedures of our proposed multi-stage training framework. As shown in Figure \ref{figure:3}, it totally consists of four stages: 1) pretraining the text encoder and text decoder on a large-scale bilingual corpus; 2) pretraining the multimodal codebook on a large-scale monolingual corpus; 3) using additional OCR data to train the image encoder and multimodal codebook via an image-text alignment task; 4) finetuning the whole model on our released TIT dataset.

\textbf{Stage 1.} We first pretrain the text encoder and text decoder on a large-scale bilingual corpus $D_{bc}$ in the way of a vanilla machine translation. Formally, for each parallel sentence $(\mathbf{x}, \mathbf{y}){\in}D_{bc}$, we define the following training objective for this stage:
{
\begin{equation}
    \mathcal{L}_{1}(\bm{\theta}_{te},\bm{\theta}_{td})=-\sum_{t=1}^{|\mathbf{y}|}\mathrm{log}(p(y_t|\mathbf{x},\mathbf{y}_{<t})),
\end{equation}
}where $\bm{\theta}_{te}$ and $\bm{\theta}_{td}$ denote the trainable parameters of the \textbf{t}ext \textbf{e}ncoder and \textbf{t}ext \textbf{d}ecoder, respectively.

\textbf{Stage 2.} This stage serves as an intermediate phase, where we exploit monolingual data to pretrain the multimodal codebook. Through this stage of training, we will learn a clustering representation for each latent code of the multimodal codebook. 

Concretely, we utilize the same dataset as the first stage but only use its source texts. Following \citeauthor{DBLP:conf/nips/OordVK17} (\citeyear{DBLP:conf/nips/OordVK17}), we update the multimodal codebook with an exponential moving average (EMA), where a decay factor determines the degree to which past values affect the current average. Formally, the latent code embedding $e_k$ is updated as follows:
{
\begin{equation}
    \begin{split}
    c_k &= \sum_{i=1}^{N_e}\mathbb{I}
    (z_q(h_{e,i}^{(L_e)})=e_k), \\
    h_k &= \sum_{i=1}^{N_e}\mathbb{I}
    (z_q(h_{e,i}^{(L_e)})=e_k)h_{e,i}^{(L_e)}, \\
    n_k &\leftarrow \gamma n_k + (1-\gamma)c_k, \\
    e_k &\leftarrow \frac{1}{n_k}(\gamma e_k + (1-\gamma)h_k),
    \label{codebook_update}
    \end{split}
\end{equation}
% \begin{align}
% \autobreak
%     c_k &= \sum_{i=1}^{N_e}\mathbb{I}
%     (z_q(h_{e,i}^{(L_e)})=e_k), \\
%     h_k &= \sum_{i=1}^{N_e}\mathbb{I}
%     (z_q(h_{e,i}^{(L_e)})=e_k)h_{e,i}^{(L_e)}, \\
%     n_k &\leftarrow \gamma n_k + (1-\gamma)c_k, \\
%     e_k &\leftarrow \frac{1}{n_k}(\gamma e_k + (1-\gamma)h_k),
%     \label{codebook_update}
% \end{align}
}\noindent where $\mathbb{I}(\cdot)$ is the indicator function and $\gamma$ is a decay factor we set to 0.99, as implemented in \cite{DBLP:conf/nips/OordVK17}. $c_k$ counts the number of text encoder hidden states that are clustered into the $k$-th latent code, $h_k$ denotes the sum of these hidden states, and $n_k$ represents the sum of the past exponentially weighted average and the current value $c_k$. Particularly, $n_k$ is set to 0 at the beginning.

\textbf{Stage 3.} During this stage, we introduce an image-text alignment task involving an additional OCR dataset $D_{ocr}$ to further train the image encoder and multimodal codebook. Through this stage of training, we expect to endow the multimodal codebook with the preliminary capability of associating images with related texts. 

Given an image-text training instance $(\mathbf{v}, \mathbf{x})\in D_{ocr}$, we define the training objective at this stage as
{
\begin{gather}
    \mathcal{L}_{3} = \mathcal{L}_{ita} + \alpha \mathcal{L}_{ic}, \label{eq:l3} \\
    \mathcal{L}_{ita}(\bm{\theta}_{ie}) = ||z_{\overline{q}}(\mathbf{H}_{v}^{(L_v)})-\mathrm{sg}(z_{\overline{q}}(\mathbf{H}_{e}^{(L_e)}))||_2^2, \\
    \mathcal{L}_{ic}(\bm{\theta}_{ie}) = ||\mathbf{H}_{v}^{(L_v)}-\mathrm{sg}(z_q(\mathbf{H}_{v}^{(L_v)}))||_2^2,
\end{gather}
}where $\mathrm{sg}(\cdot)$ refers to a stop-gradient operation and $\bm{\theta}_{ie}$ is the parameters of the \textbf{i}mage \textbf{e}ncoder except the $\mathrm{ViT}$ module. Specifically, $z_{\overline{q}}(\mathbf{H}_{v}^{(L_v)})$ is calculated as $\frac{1}{N_v}\sum_{j=1}^{N_v}z_q({h}_{v,j}^{(L_v)})$ and $z_{\overline{q}}(\mathbf{H}_{e}^{(L_e)})$ is calculated as $\frac{1}{N_e}\sum_{i=1}^{N_e}z_q({h}_{e,i}^{(L_e)})$, which represent the semantic information of image and text respectively. Via $\mathcal{L}_{ita}$, we expect to enable both image and text representations to be quantized into the same latent codes. Meanwhile, following \citeauthor{DBLP:conf/nips/OordVK17} (\citeyear{DBLP:conf/nips/OordVK17}), we use the commitment loss $\mathcal{L}_{ic}$ to ensure that the output hidden states of image encoder stay close to the chosen latent code embedding, preventing it fluctuating frequently from one latent code to another, and $\alpha$ is a hyperparameter to control the effect of $\mathcal{L}_{ic}$. Note that at this stage, we continue to update the parameters of the multimodal codebook using Equation \ref{codebook_update}.

\textbf{Stage 4.}
Finally, we use the TIT dataset $D_{tit}$ to finetune the whole model. Notably, $\mathcal{L}_{3}$ is still involved, which maintains the training consistency and makes finetuning smoothing.

Given a TIT training instance $(\mathbf{v},\hat{\mathbf{x}},\mathbf{x},\mathbf{y}){\in}D_{tit}$, we optimize the whole model through the following objective:
{
\begin{gather}
    \mathcal{L}_{4}=\mathcal{L}_{3} + \mathcal{L}_{tit} + \beta \mathcal{L}_{tc}, \label{eq:l4}\\
    \mathcal{L}_{tit}(\bm{\theta}_{te},\bm{\theta}_{ie},\bm{\theta}_{td})=-\sum_{t=1}^{|\mathbf{y}|}\mathrm{log}(p(y_t|\mathbf{v},\hat{\mathbf{x}},\mathbf{y}_{<t})),\\
    \mathcal{L}_{tc}(\bm{\theta}_{te}) = ||\mathbf{H}_{e}^{(L_e)}-\mathrm{sg}(z_q(\mathbf{H}_{e}^{(L_e)}))||_2^2, 
\end{gather}
}\noindent where $\mathcal{L}_{tc}$ is also a commitment loss proposed for the text encoder, and $\beta$ is a hyperparameter quantifying its effect. Note that $\hat{\mathbf{x}}$ is only used as an input for $\mathcal{L}_{tit}$ to ensure the consistency between the model training and inference, and $\mathbf{x}$ is used as an input for image-text alignment task to train the ability of the multimodal codebook in associating the input image with the ground-truth text. Besides, we still update the multimodal codebook with EMA.

\section{Experiments}
\subsection{Datasets}
Our proposed training framework consists of four stages, involving the following three datasets:

\textbf{WMT22 ZH-EN}\footnote{https://www.statmt.org/wmt22/translation-task.html}. This large-scale parallel corpus contains about 28M parallel sentence pairs and we sample 2M parallel sentence pairs from the original whole corpus. During the first and second training stages, we use the sampled dataset to pretrain our text encoder and text decoder.

\textbf{ICDAR19-LSVT}. It is an OCR dataset including 450, 000 images with texts that are freely captured in the streets, e.g., storefronts and landmarks. In this dataset, 50,000 fully-annotated images are partially selected to construct the OCRMT30K dataset, and the remaining 400,000 images are weakly annotated, where only the text-of-interest in these images are provided as ground truths without location annotations. In the third training stage, we use these weakly annotated data to train the image encoder and multimodal codebook via the image-text alignment task.

\textbf{OCRMT30K}. As mentioned previously, our OCRMT30K dataset involves five Chinese OCR datasets: RCTW-17, CASIA-10K, ICDAR19-MLT, ICDAR19-LSVT, and ICDAR19-ArT. It totally contains about 30,000 instances, where each instance involves an image paired with several Chinese texts and their corresponding English translations. In the experiments, we choose 1,000 instances for development, 1,000 for evaluation, and the remaining instances for training. Besides, We use the commonly-used PaddleOCR to handle our dataset and obtain the recognized texts. In the final training stage, we use the training set of OCRMT30K to finetune our whole model.

\subsection{Settings}
We use the standard ViT-B/16 \cite{DBLP:conf/iclr/DosovitskiyB0WZ21} to model our image encoder. Both our text encoder and text decoder consist of 6 layers, each of which has 512-dimensional hidden sizes, 8 attention heads, and 2,048 feed-forward hidden units. Particularly, a 512-dimensional word embedding layer is shared across the text encoder and the text decoder. We set the size of the multimodal codebook to 2,048.

During the third stage, following \citeauthor{DBLP:conf/nips/OordVK17} (\citeyear{DBLP:conf/nips/OordVK17}), we set $\alpha$ in Equation \ref{eq:l3} to 0.25. During the final training stage, we set $\alpha$ to 0.75 and $\beta$ in Equation \ref{eq:l4} to 0.25 determined by a grid search on the validation set, both of which are varied from 0.25 to 1 with an interval of 0.25. We use the batch size of 32,768 tokens in the first and second training stages and 4,096 tokens in the third and final training stages. In all stages, we apply the Adam optimizer \cite{DBLP:journals/corr/KingmaB14} with $\beta_1$ = 0.9, $\beta_2$ = 0.98 to train the model, where the inverse square root schedule algorithm and warmup strategy are adopted for the learning rate. Besides, we set the dropout to 0.1 in the first three training stages and 0.3 in the final training stage, and the value of label smoothing to 0.1 in all stages.

During inference, we use beam search with a beam size of 5. Finally, we employ BLEU \cite{DBLP:conf/acl/PapineniRWZ02} calculated by SacreBLEU\footnote{https://github.com/mjpost/sacrebleu} \cite{DBLP:conf/wmt/Post18} and COMET\footnote{https://github.com/Unbabel/COMET} \cite{DBLP:conf/emnlp/ReiSFL20} to evaluate the model performance.

\subsection{Baselines}
In addition to the text-only Transformer \cite{DBLP:conf/nips/VaswaniSPUJGKP17}, our baselines include:

\begin{itemize}
\item \textit{Doubly-ATT} \cite{DBLP:conf/acl/CalixtoLC17}. This model uses two attention mechanisms to exploit the image and text representations for translation, respectively.  
\item \textit{Imagination} \cite{DBLP:conf/ijcnlp/ElliottK17}. It trains an image prediction decoder to predict a global visual feature vector that is associated with the input sentence.
\item \textit{Gated Fusion} \cite{DBLP:conf/acl/WuKBLK20}. This model uses a gated vector to fuse image and text representations, and then feeds them to a decoder for translation. 
\item \textit{Selective Attn} \cite{DBLP:conf/acl/LiLZZXMZ22}. It is similar to \textit{Gated Fusion}, but uses a selective attention mechanism to make better use of the patch-level image representation. 
\item \textit{VALHALLA} \cite{DBLP:conf/cvpr/LiPKCFCV22}. This model uses an autoregressive hallucination Transformer to predict discrete visual representations from the input text, which are then combined with text representations to obtain the target translation.
% \item The retrieval-based MMT models: \textit{UVR-NMT} \cite{DBLP:conf/iclr/uvrnmt} build a token-image lookup table from an image-text dataset and then retrieve images matching the source keywords to help subsequent translations.
\item \textit{E2E-TIT} \cite{DBLP:conf/icpr/MaZTHWZ022}. It applies a multi-task learning framework to train an end-to-end TIT model, where MT and OCR serve as auxiliary tasks.  Note that except for E2E-TIT, all other models are cascaded ones. Unlike other cascaded models that take recognized text and the entire image as input, the input to this end-to-end model is an image cropped from the text bounding box.
\end{itemize}

To ensure fair comparisons, we pretrain all these baselines on the same large-scale bilingual corpus.

\begin{table}[]
\centering
\setlength\tabcolsep{3pt}
\resizebox{\columnwidth}{!}{
% \tiny
\begin{tabular}{cccc}
\hline
\multicolumn{1}{l|}{\textbf{Model}} & \textbf{BLEU}  & \textbf{COMET}                   \\  \hline
\multicolumn{3}{c}{\textit{Text-only Transformer}}                                        \\ \hline
\multicolumn{1}{l|}{Transformer \cite{DBLP:conf/nips/VaswaniSPUJGKP17}}                     &39.38                          &30.01       \\ \hline
\multicolumn{3}{c}{\textit{Existing MMT Systems}}                                         \\ \hline
\multicolumn{1}{l|}{Imagination \cite{DBLP:conf/ijcnlp/ElliottK17}}                      &39.47                              &30.66    \\ 
\multicolumn{1}{l|}{Doubly-ATT \cite{DBLP:conf/acl/CalixtoLC17}}                      &39.93                             &30.52     \\ 
\multicolumn{1}{l|}{Gated Fusion \cite{DBLP:conf/acl/WuKBLK20}}                      &40.03                            &30.91       \\ 
\multicolumn{1}{l|}{Selective Attn \cite{DBLP:conf/acl/LiLZZXMZ22}}                      &39.82                            &30.82       \\ 
% \multicolumn{3}{c}{\textit{Image Association Systems}}                                         \\ \hline
% \multicolumn{1}{l|}{Imagination \cite{DBLP:conf/ijcnlp/ElliottK17}}                      &                 &            &      \\ 
\multicolumn{1}{l|}{VALHALLA \cite{DBLP:conf/cvpr/LiPKCFCV22}}                      &39.73                          &30.10        \\ \hline
\multicolumn{3}{c}{\textit{Existing TIT System}}                                         \\ \hline
\multicolumn{1}{l|}{E2E-TIT \cite{DBLP:conf/icpr/MaZTHWZ022}}                         &19.50                         &-31.90          \\ \hline
\multicolumn{3}{c}{\textit{Our TIT System}}                                               \\ \hline
\multicolumn{1}{l|}{Our model}                                &$\textbf{40.78}^{\ddagger}$                           &$\textbf{33.09}^{\ddagger}$         \\ \hline
\end{tabular}
}
\caption{Experimental results on the Zh$\rightarrow$En TIT task. “$\ddagger$” represents the improvement over the best result of all other contrast models is statistically significant ($p{<}0.01$).}
\label{tab:main result}
\end{table}

\subsection{Results}
Table \ref{tab:main result} reports the performance of all models. We can observe that our model outperforms all baselines, achieving state-of-the-art results. Moreover, we draw the following interesting conclusions:

\textit{First}, all cascaded models exhibit better performance than E2E-TIT. For this result, we speculate that as an end-to-end model, E2E-TIT may struggle to distinguish text from the surrounding background in the image when the background exhibits visual characteristics similar to the text.

\textit{Second}, our model outperforms Doubly-ATT, Gated Fusion, and Selective Attn, all of which adopt attention mechanisms to exploit image information for translation. The underlying reason is that each input image and its texts are mapped into the shared semantic space of latent codes, reducing the modality gap and thus enabling the model to effectively utilize image information.

\textit{Third}, our model also surpasses Imagination and VALHALLA, both of which use the input text to generate the representations of related images. We conjecture that in the TIT task, it may be challenging for the model to generate useful image representations from the incorrectly recognized text. In contrast, our model utilizes the input image to generate related text representations, which is more suitable for the TIT task.

 Inspired by E2E-TIT, we also compare other baselines with the cropped image as input. Table \ref{tab:appendix_result} reports the results of our model compared with other baselines using the cropped image as input. We can observe that our model still achieves state-of-the-art results.

\begin{table}[]
\centering
\setlength\tabcolsep{3pt}
\resizebox{\columnwidth}{!}{
% \tiny
\begin{tabular}{cccc}
\hline
\multicolumn{1}{l|}{\textbf{Model}} & \textbf{BLEU}  & \textbf{COMET}                   \\  \hline
\multicolumn{3}{c}{\textit{Text-only Transformer}}                                        \\ \hline
\multicolumn{1}{l|}{Transformer \cite{DBLP:conf/nips/VaswaniSPUJGKP17}}                     &39.38                          &30.01       \\ \hline
\multicolumn{3}{c}{\textit{Existing MMT Systems}}                                         \\ \hline
\multicolumn{1}{l|}{Imagination \cite{DBLP:conf/ijcnlp/ElliottK17}}                      &39.64                              &30.68    \\ 
\multicolumn{1}{l|}{Doubly-ATT \cite{DBLP:conf/acl/CalixtoLC17}}                      &39.71                             &31.42     \\ 
\multicolumn{1}{l|}{Gated Fusion \cite{DBLP:conf/acl/WuKBLK20}}                      &39.03                            &30.46       \\ 
\multicolumn{1}{l|}{Selective Attn \cite{DBLP:conf/acl/LiLZZXMZ22}}                      &40.13                            &30.74       \\ 
% \multicolumn{3}{c}{\textit{Image Association Systems}}                                         \\ \hline
% \multicolumn{1}{l|}{Imagination \cite{DBLP:conf/ijcnlp/ElliottK17}}                      &                 &            &      \\ 
\multicolumn{1}{l|}{VALHALLA \cite{DBLP:conf/cvpr/LiPKCFCV22}}                      &39.24                          &29.08        \\ \hline
\multicolumn{3}{c}{\textit{Existing TIT System}}                                         \\ \hline
\multicolumn{1}{l|}{E2E-TIT \cite{DBLP:conf/icpr/MaZTHWZ022}}                         &19.50                         &-31.90          \\ \hline
\multicolumn{3}{c}{\textit{Our TIT System}}                                               \\ \hline
\multicolumn{1}{l|}{Our model}                                &$\textbf{40.78}^{\ddagger}$                           &$\textbf{33.09}^{\dagger}$         \\ \hline
\end{tabular}
}
\caption{Additional experimental results on the Zh$\rightarrow$En TIT task. “$\ddagger/\dagger$” represents the improvement over the best result of all other contrast models is statistically significant ($p{<}0.01/0.05$).}
\label{tab:appendix_result}
\end{table}

\begin{figure*}[t]
\setlength{\abovecaptionskip}{2pt}
    \centering
    \includegraphics[width=0.85\linewidth]{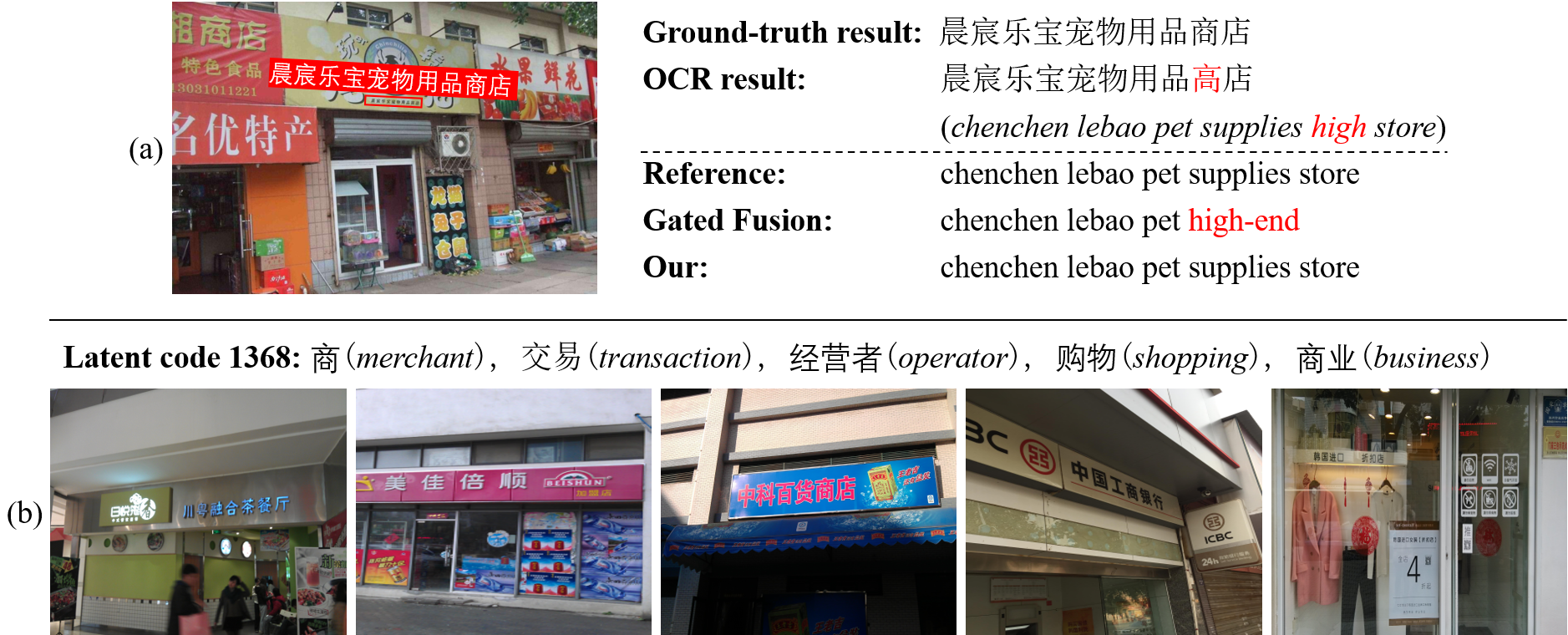}
    \caption{An example of TIT task on the OCRMT30K dataset.}
    \label{fig:visualize}
\end{figure*}

% \vspace{-0.1cm}
\subsection{Ablation Study}
To investigate the effectiveness of different stages and modules, we further compare our model with several variants in Table \ref{tab:ablation study}:

\textit{w/o Stage 2}. We remove the second training stage in this variant. The result in line 2 shows that this change causes a significant performance decline. It suggests that pretraining the clustering representations of latent codes in the multimodal codebook is indeed helpful for the model training. 

\textit{w/o Stage 3}. In this variant, we remove the third stage of training. The result in line 3 indicates that this removal leads to a performance drop. The result confirms our previous assumption that training the preliminary capability of associating images and related texts indeed enhances the TIT model.

\textit{w/o $\mathcal{L}_{3}$ in Stage 4}. When constructing this variant, we remove the loss item $\mathcal{L}_{3}$ from stage 4. From line 4, we can observe that preserving $\mathcal{L}_{3}$ in the fourth stage makes the transition from the third to the fourth stage smoother, which further alleviates the training discrepancy.

\textit{w/o multimodal codebook}. We remove the multimodal codebook in this variant, and the visual features extracted through the image encoder are utilized in its place. Apparently, the performance drop drastically as reported in line 5, demonstrating the effectiveness of the multimodal codebook. 

\textit{w/ randomly sampling latent codes}. Instead of employing quantization, we randomly sample latent codes from the multimodal codebook in this variant. Line 6 shows that such sampling leads to a substantial performance decline. Thus, we confirm that latent codes generated from the input image indeed benefits the subquent translation.

\begin{table}[]
\centering
\resizebox{\columnwidth}{!}{
\begin{tabular}{l|cc}
\hline
\textbf{Model}          & \textbf{BLEU} & \textbf{COMET} \\ \hline
Our model               &\textbf{40.78}               &\textbf{33.09}                \\ \hline
\hspace{1em}w/o Stage 2              &39.93               &31.35                \\
\hspace{1em}w/o Stage 3              &40.15               &30.90                \\
\hspace{1em}w/o $\mathcal{L}_{3}$ in Stage 4        &40.18               &31.99                \\
\hspace{1em}w/o multimodal codebook &38.81               &29.08                \\ 
\hspace{1em}w/ randomly sampling latent codes &34.91               &18.90                \\ \hline
\end{tabular}
}
\caption{Ablation study of our model on the Zh$\rightarrow$En text image translation task.}
\label{tab:ablation study}
\end{table}

\subsection{Analysis}

To further reveal the effect of the multimodal book, we provide a translation example in Figure \ref{fig:visualize}(a), listing the OCR result and translations produced by ours and Gated Fusion, which is the most competitive baseline.
It can be seen that \begin{CJK}{UTF8}{gbsn}“\textit{用品商店}”\end{CJK} (“\textit{supplies store}”) is incorrectly recognized as \begin{CJK}{UTF8}{gbsn}“\textit{用品高店}”\end{CJK} 
 (“\textit{supplies high store}”), resulting in the incorrect translation even for Gated Fusion.
By contrast, our model can output the correct translation with the help of the multimodal codebook.

During decoding for “supplies store”, latent code 1368 demonstrated the highest cross-attention weight in comparison to other codes. Therefore, we only visualize the latent code 1368 for analysis. In Figure \ref{fig:visualize}(b), since tokens may be duplicated and all images are different, we provide the five most frequent tokens and five randomly-selected images from this latent code, and find that all these tokens and images are highly related to the topic of business. Thus, intuitively, the clustering vector of this latent code will fully encode the information related to the business, and thus can provide useful information to help the model conduct the correct translation. 

\section{Conclusion}
In this paper, we release a Chinese-English TIT dataset named OCRMT30K, which is the first publicly available TIT dataset. Then, we propose a novel TIT model with a multimodal codebook. Typically, our model can leverage the input image to predict latent codes associated with the input sentence via the multimodal codebook, providing supplementary information for the subsequent translation. Moreover, we present a multi-stage training framework that effectively utilizes additional bilingual texts and OCR data to refine the training of our model. 

In the future, we intend to construct a larger dataset and explore the potential applications of our method in other multimodal tasks, such as video-guided machine translation.

\section*{Limitations}
Since our model involves an additional step of OCR, it is less efficient than the end-to-end TIT model, although it can achieve significantly better performance. Besides, with the incorporation of image information, our model is still unable to completely address the issue of error propagation caused by OCR.

\section*{Ethics Statement}
This paper proposes a TIT model and a multi-stage training framework. We take ethical considerations seriously and ensure that the methods used in this study are conducted in a responsible and ethical manner. We also release a Chinese-English TIT dataset named OCRMT30K, which is annotated based on five publicly available Chinese OCR datasets, and are used to support scholars in doing research and not for commercial use, thus there exists not any ethical concern.

\section*{Acknowledgments}
The project was supported by National Key Research and Development Program of China (No. 2020AAA0108004), National Natural Science Foundation of China (No. 62276219), and Natural Science Foundation of Fujian Province of China (No. 2020J06001). We also thank the reviewers for their insightful comments.

% Entries for the entire Anthology, followed by custom entries
\bibliography{anthology,custom}
\bibliographystyle{acl_natbib}

% \appendix
% \section{Appendix}
% \section{Additional Results}
% \label{appendix:addresult}

\end{document}